\pdfoutput=1

\documentclass[11pt]{article}

\usepackage{coling}
\usepackage[symbol]{footmisc}
\usepackage{times}
\usepackage{latexsym}
\usepackage{adjustbox}
\usepackage{placeins}
\usepackage{enumitem}
\setlist{nosep} 
\usepackage[T1]{fontenc}

\usepackage[utf8]{inputenc}

\usepackage{microtype}

\usepackage{inconsolata}

\usepackage{graphicx}
\usepackage{tikz}

\usepackage{times}
\usepackage{latexsym}

\usepackage{times}
\usepackage{latexsym}
\usepackage{amsmath}
\usepackage{amssymb}
\usepackage{tabularx}
\usepackage{adjustbox}
\usepackage{multirow}
\usepackage{booktabs}
\usepackage{float}
\usepackage{stfloats}
\usepackage{pgfplots}
\pgfplotsset{compat=1.17}
\usepackage{xcolor}
\usepackage{colortbl}

\definecolor{COT}{HTML}{e68193}    
\definecolor{SC}{HTML}{9e9241} 
\definecolor{TREE}{HTML}{43a178}  
\definecolor{TOT}{HTML}{4a9eb4} 
\definecolor{STANDARD}{HTML}{c789e5}  
\definecolor{S2A}{HTML}{0244b6} 
\definecolor{PAL}{HTML}{ff6d01}

\title{Can We Afford The Perfect Prompt? Balancing Cost and Accuracy with the \textsc{Economical Prompting Index}}

\author{
    Tyler McDonald\thanks{Equal contribution.}\textnormal{,}
    Anthony Colosimo\footnotemark[1]\textnormal{,} 
    Yifeng Li\textnormal{, and}
    Ali Emami\\
    Brock University, Saint Catharines, Canada \\
    \texttt{\{tmcdonald3, ac19fa, yli2, aemami\}@brocku.ca} \\
}

\begin{document}
\maketitle
\begin{abstract}
As prompt engineering research rapidly evolves, evaluations beyond accuracy are crucial for developing cost-effective techniques. We present the Economical Prompting Index (EPI), a novel metric that combines accuracy scores with token consumption, adjusted by a user-specified cost concern level to reflect different resource constraints. Our study examines 6 advanced prompting techniques, including Chain-of-Thought, Self-Consistency, and Tree of Thoughts, across 10 widely-used language models and 4 diverse datasets. We demonstrate that approaches such as Self-Consistency often provide statistically insignificant gains while becoming cost-prohibitive. For example, on high-performing models like Claude 3.5 Sonnet, the EPI of simpler techniques like Chain-of-Thought (0.72) surpasses more complex methods like Self-Consistency (0.64) at slight cost concern levels. Our findings suggest a reevaluation of complex prompting strategies in resource-constrained scenarios, potentially reshaping future research priorities and improving cost-effectiveness for end-users.
\end{abstract}

\section{Introduction}

Prompt engineering is a growing subdiscipline of natural language processing, providing a consumer-friendly alternative to fine-tuning methods. Recent research focuses on enhancing reasoning in Large Language Models (LLMs) across various domains of problem-solving, such as arithmetic, commonsense, spatial, and multimodal reasoning \citep{wei2023chainofthought, yao2023tree, gu2023systematic, ma2023prompt}.

With many new works being published in rapid succession, there has been an increased reliance on accuracy gains as the primary justification for new techniques \citep{bender-koller-2020-climbing, lipton2018troublingtrendsmachinelearning}. Though iterative and recursive techniques promise accuracy improvements through deliberate and continuous reasoning, there is a lack of appropriate consideration for the financial constraints of computationally burdensome methods \citep{sahoo2024systematic}. This oversight can lead to techniques that, while marginally more accurate, may be prohibitively expensive for practical applications, potentially limiting their adoption and real-world impact. Token usage serves as an effective proxy for computational cost, as it directly correlates with the resources required for model inference and often forms the basis for pricing in commercial LLM services.

\begin{figure}[t]
    \centering
    \includegraphics[width=\columnwidth]{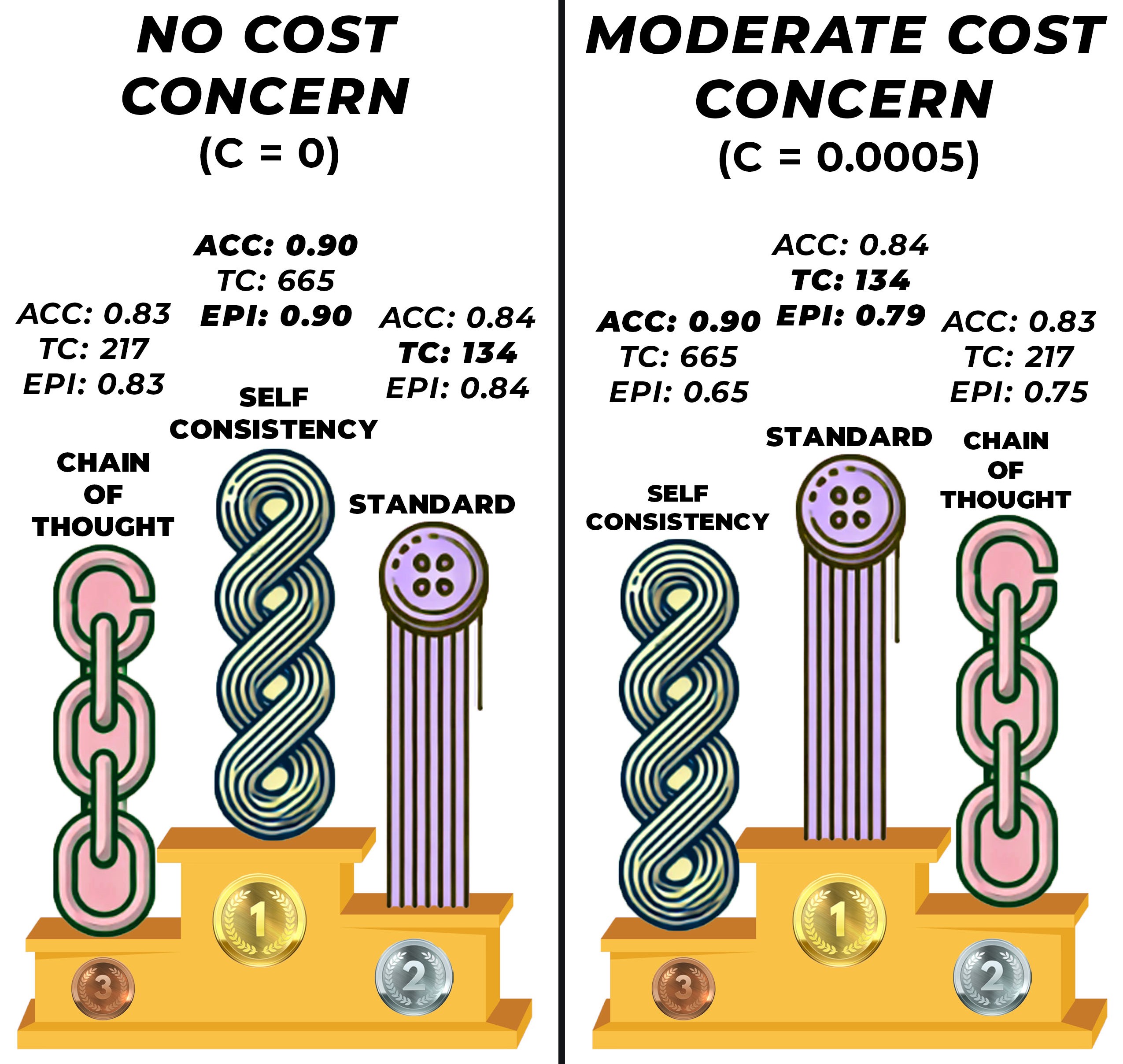}
    
    \caption{Economical Prompting Index (EPI) for GPT-4 across datasets, comparing no cost concern (C = 0) and moderate cost concern (C = 0.0005) scenarios. Prompt rankings shift when considering both accuracy (ACC) and token cost (TC).}
    \label{fig:cover}
\end{figure}

To address this gap, we propose evaluating popular techniques through dimensions beyond accuracy alone. Our approach aims to provide a more holistic assessment of prompting techniques, discouraging the development of new methods solely for the purpose of incremental gains. To demonstrate this, we introduce the \textsc{Economical Prompting Index} (EPI), a novel metric for evaluating prompting techniques that considers both token usage alongside dataset accuracy.

Figure \ref{fig:cover} illustrates the utility of the EPI, showcasing results obtained from GPT-4 averaged across all datasets studied. The EPI demonstrates how different levels of cost concern can significantly alter the perceived effectiveness of various prompting techniques. For instance, while Self-Consistency shows high efficacy at low-cost concern levels, its effectiveness diminishes rapidly as cost considerations increase, with simpler methods like Chain-of-Thought becoming more favorable.

Building on these insights, our work makes several key contributions. We introduce the \textsc{Economical Prompting Index} (EPI), a novel metric that balances accuracy with resource usage, providing a flexible, user-centric measure of prompting methods' efficacy. Our comprehensive evaluation of 6 prompting techniques across 4 diverse datasets and 10 flagship language models demonstrates the varying performance and resource implications of different methods. By applying the EPI to our experimental results, we reveal how the relative effectiveness of techniques like Self-Consistency can shift dramatically when resource utilization is considered, often favoring simpler, more cost-effective methods in practical scenarios.\footnote{The complete code and detailed interactions with the language models can be found \href{https://github.com/tm21cy/EconomicalPromptingIndex}{here}.}

\section{Economical Prompting Index}

The \textsc{Economical Prompting Index} (EPI) addresses the need for a comprehensive metric that balances accuracy with token consumption in prompt design, providing a more complete evaluation of prompting techniques.

\begin{table}[t]
\small
\centering
\begin{tabular}{@{}p{1.3cm}p{1cm}l@{}}
\hline
\textbf{Cost \newline Concern} & \textbf{$C$ Value} & \textbf{Application Scenario} \rule{0pt}{2.25ex}    \\
\hline
None & 0 & Research with unlimited resources \\
Slight & 0.00025 & Production with ample resources \\
Moderate & 0.0005 & Typical commercial applications \\
Elevated & 0.001 & Resource-constrained environments \\
Major & 0.002 & Highly cost-sensitive scenarios \\
\hline
\end{tabular}
\vspace{-1mm}
\caption{Weight classes for cost concern factor $C$.}
\label{tab:cost-concern}
\end{table}

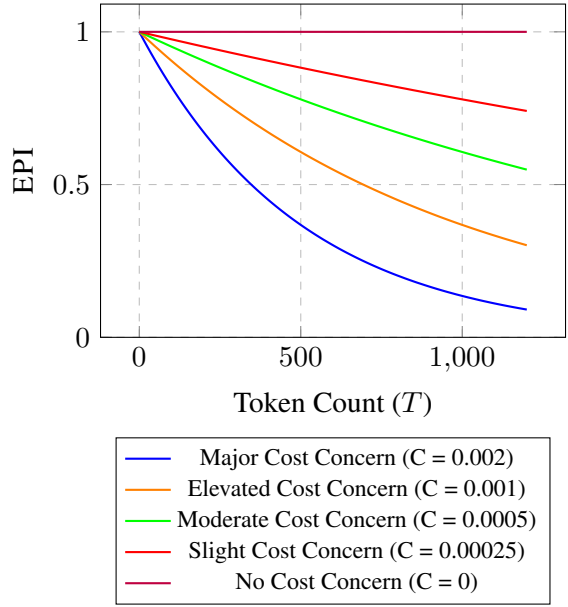
\begin{figure}[htbp] 
    \centering
    \begin{tikzpicture}
    \begin{axis}[
        xlabel={Token Count ($T$)},
        ylabel={\textsc{EPI}},
        legend style={
            at={(0.5,-0.3)}, 
            anchor=north, 
            legend columns=1, 
            font=\small, 
            /tikz/every even column/.append style={column sep=0.5cm}
        },
        grid style=dashed,
        grid=both,
        width=\columnwidth, 
        height=6cm,
        every axis plot/.append style={thick},
        cycle list name=color list
    ]

    \def\Y{1}

    \pgfmathdeclarefunction{Yadjusted}{3}{%
        \pgfmathparse{#1 * exp(-#3 * #2)}%
    }

    \addplot [
        domain=0:1200, 
        samples=100,
        blue,
        solid
    ] {Yadjusted(\Y, x, 0.002)};
    \addlegendentry{Major Cost Concern (C = 0.002)}

    \addplot [
        domain=0:1200, 
        samples=100,
        orange,
        solid
    ] {Yadjusted(\Y, x, 0.001)};
    \addlegendentry{Elevated Cost Concern (C = 0.001)}

    \addplot [
        domain=0:1200, 
        samples=100,
        green,
        solid
    ] {Yadjusted(\Y, x, 0.0005)};
    \addlegendentry{Moderate Cost Concern (C = 0.0005)}

    \addplot [
        domain=0:1200, 
        samples=100,
        red,
        solid
    ] {Yadjusted(\Y, x, 0.00025)};
    \addlegendentry{Slight Cost Concern (C = 0.00025)}

    \addplot [
        domain=0:1200, 
        samples=100,
        purple,
        solid
    ] {Yadjusted(\Y, x, 0)};
    \addlegendentry{No Cost Concern (C = 0)}

    \end{axis}
    \end{tikzpicture}
    \vspace{-1mm}
    \caption{Graph of the token count $T$ against the \textsc{EPI} given the demonstrative weight classes $C$, for $A=1$.}
    \label{fig:plot}
\end{figure}

For any prompting technique $\mathcal{P}$ and input question $Q$, we obtain a response with accuracy $A$ and a total token count $T$ (both input and output tokens):
\begin{gather*}
\mathcal{P}({Q}) \to {A}, {T}
\end{gather*}

\noindent where $A \in [0,1]$ is the proportion of correct responses and $T \in \mathbb{N}$ is the total number of tokens used.
The EPI $\in [0,1]$ calculates a final performance measure by including a cost concern factor $C \in [0,1]$:
$$
EPI({A,C,T}) = A \times e^{ (-C \times T)}
$$

The cost concern factor $C$ represents the relative importance of resource efficiency in a given application scenario, with higher values indicating greater sensitivity to token usage.

Alternative considerations, such as linear and polynomial functions, were explored as per Appendix \ref{sec:epi-appendix}; however, linear models fail to adequately capture the cost-performance tradeoff in sensitive settings (for $C=0.00083$, $EPI=0.7925$ at $T=250$ and $EPI=0.585$ at $T=500$ --- a 26\% reduction despite requiring double the tokens), while polynomial approaches exhibit vanishing behavior at large values of $T$ (for $C=1.5\times10^{-6}$ and $T>666$, $EPI=0$). As such, an exponential model was chosen to faithfully depict the concerns associated with mounting token cost at various sensitivities.

Interpreting the EPI:
\begin{itemize}[itemsep=0pt, leftmargin=*]
\setlength{\itemsep}{-2pt} 
\setlength{\parsep}{-2pt}  
\item When $C = 0$, EPI equals the raw accuracy ($A$), providing a baseline for cost-aware scenarios.
\item As EPI approaches 1, it indicates high accuracy with efficient token usage relative to the chosen cost concern level.
\item A low EPI suggests either poor accuracy, excessive token usage, or both, depending on the specific values of $A$, $C$, and $T$.
\item For a given $C>0$ , techniques with similar EPI values represent comparable trade-offs between accuracy and efficiency, even if their raw accuracy and token counts differ.
\end{itemize}
We provide five representative weight classes sampled from the continuous range of $C$, shown in Table 
\ref{tab:cost-concern}. Figure \ref{fig:plot} shows how the EPI changes with token count for different levels of cost concern, assuming perfect accuracy ($A=1$).



\begin{table*}[t]
\small
\centering
\begin{adjustbox}{width=\textwidth}
\begin{tabular}{lcccccccc}
\toprule
\textbf{Prompt Methods} & \multicolumn{2}{c}{\textbf{CSQA}} & \multicolumn{2}{c}{\textbf{MMLU}} & \multicolumn{2}{c}{\textbf{GSM8K}} & \multicolumn{2}{c}{\textbf{DQA}} \\
\cmidrule(lr){2-3} \cmidrule(lr){4-5} \cmidrule(lr){6-7} \cmidrule(lr){8-9}
 & \textbf{Accuracy} & \textbf{Token Count} & \textbf{Accuracy} & \textbf{Token Count} & \textbf{Accuracy} & \textbf{Token Count} & \textbf{Accuracy} & \textbf{Token Count} \\
\midrule
Chain-of-Thought     & 0.79 & 205.22 & 0.74 & 301.84 & 0.89 & 257.03 & 0.60 & 229.80 \\
Self-Consistency     & \textbf{\textcolor{green!60!black}{0.88}} & \phantom{*}\textbf{\textcolor{red}{619.93}}* & \textbf{\textcolor{green!60!black}{0.84}} & \phantom{*}\textbf{\textcolor{red}{902.89}}* & \textbf{\textcolor{green!60!black}{0.95}} & \phantom{*}\textbf{\textcolor{red}{773.03}}* & \textbf{\textcolor{green!60!black}{0.76}} & \phantom{*}\textbf{\textcolor{red}{689.78}}*\\
Tree of Thoughts     & 0.74 & 383.69 & 0.66 & 427.24 & 0.79 & 375.44 & 0.60 & 385.17 \\
Thread of Thought    & 0.78 & 324.17 & 0.73 & 417.74 & 0.89 & 348.68 & 0.60 & 274.41 \\
Standard             & 0.77 & \textbf{\textcolor{green!60!black}{140.77}} & 0.75 & \textbf{\textcolor{green!60!black}{221.26}} & 0.86 & \textbf{\textcolor{green!60!black}{217.95}} & 0.58 & \textbf{\textcolor{green!60!black}{161.80}} \\
System 2 Attention   & \textbf{\textcolor{red}{0.67}} & 303.55 & \textbf{\textcolor{red}{0.62}} & 401.76 & \textbf{\textcolor{red}{0.68}} & 353.76 & \textbf{\textcolor{red}{0.45}} & 363.93 \\
\bottomrule
\end{tabular}
\end{adjustbox}
\centering
\vspace{-1mm}
\caption{Accuracy and token count averaged across all models. \textbf{\textcolor{red}{Red}}: highest cost/lowest accuracy; \textbf{\textcolor{green!60!black}{green}}: lowest cost/highest accuracy. * indicates statistical significance at $p<0.05$ as outlined by our procedures in Appendix \ref{Stat-Sig}.}
\label{tab:all-dataset}
\end{table*}

\section{Experimental Setup}

\subsection{Datasets and Models}
We sampled from four diverse datasets: Grade School Math 8K (GSM8K), CommonsenseQA (CSQA), Massive Multitask Language Understanding (MMLU), and BIG-Bench Hard Disambiguation QA (DQA) \citep{cobbe2021training, talmor2019commonsenseqa, hendrycks2021measuringmassivemultitasklanguage,suzgun2022challengingbigbenchtaskschainofthought}. For each dataset, we used $n=200$ samples, except for MMLU, where we sampled 4 entries from each of its 57 subjects ($n=228$).

We evaluated 10 models from 5 publishers: OpenAI (GPT-3.5-Turbo, GPT-4), Google DeepMind (Gemini 1 Pro, Gemini 1.5 Pro), Anthropic (Claude 3 Haiku, Claude 3.5 Sonnet), Meta (Llama 3 8B, Llama 3 70B), and Mistral AI (Mixtral 8x7B, 8x22B) \citep{brown2020language, openai2024gpt4technicalreport, geminiteam2024geminifamilyhighlycapable, touvron2023llamaopenefficientfoundation, jiang2024mixtralexperts}. Llama and Mixtral models were queried via Anyscale, while others used their provided APIs.\footnote{https://www.anyscale.com/}
\subsection{Prompting Techniques and Evaluation}
We tested six prompting techniques: standard, Chain-of-Thought, Self-Consistency, Tree of Thoughts, Thread of Thought, and System 2 Attention \citep{wei2023chainofthought, wang2023selfconsistency, yao2023tree, zhou2023thread,weston20232}. All prompts were applied in a zero-shot setting (see Appendix \ref{Prompt Text} for prompt templates).
Accuracy was computed as a percentage of correct responses, and token count as the average input and output tokens per query over the full sample. 
\subsection{EPI Calculation}
We calculated \textsc{EPI} scores for each prompting technique, both 
\textit{model-specifically} (averaged across datasets per model) and \textit{model-agnostically} (averaged across models per dataset).

\section{Results}

\subsection{Accuracy and Token Count Analysis}
Table \ref{tab:all-dataset} shows the accuracy and token counts averaged across the 10 models employed for testing, while example outputs on CSQA with GPT-4 can be found in Appendix \ref{Examples}.

\noindent \textbf{Key Observations:}
\begin{itemize}[itemsep=0pt, leftmargin=*]
\item Self-Consistency provides the highest accuracy across all tasks, but at a disproportionately higher cost. On GSM8K, a statistically insignificant 6.74\% increase in performance comes with a 200\% increase in token consumption compared to standard prompting.
\item System 2 Attention, despite its complexity, shows the lowest accuracy across all datasets, suggesting that more elaborate techniques do not always yield better results.
\end{itemize}

\subsection{Model-Agnostic EPI Results}
Figure 3 illustrates the application of the EPI across different prompting techniques on GSM8K. Our analysis reveals the following key findings:

\begin{itemize}[itemsep=0pt, leftmargin=*]
    \item \textbf{Self-Consistency shows rapid deterioration in cost efficacy} (slope $m = -361.09$) as cost concern increases, indicating a steep decline in effectiveness when considering token usage. This slope represents the rate at which the EPI decreases as the cost concern factor increases.
    
    \item \textbf{Chain-of-Thought demonstrates slower deterioration} ($m = -177.22$), indicating better viability under cost constraints. The shallower slope suggests this method strongly retains its effectiveness as cost concerns grow.
\end{itemize}

Additional visualizations can be found in Appendix Section \ref{Model-Agnostic}.

\subsection{Model-Specific EPI Results}
Figure \ref{fig:claude3.5-line} shows the EPI scores for Claude 3.5 Sonnet across all datasets. Analysis of these results yields the following insights:

\begin{itemize}[itemsep=0pt, leftmargin=*]

\item \textbf{For high-performing models like Claude 3.5 Sonnet, complex techniques offer only incremental gains} (e.g. Self-Consistency: 0.83, Chain-of-Thought: 0.79). Moreover, these gains are not statistically significant on all tasks except MMLU ($p < 0.05$).
    
\item \textbf{As cost concern increases, simpler techniques like Chain-of-Thought become more viable} (overtaking Self-Consistency at $C=0.00008$). This intersection point indicates that even at very low levels of cost concern, simpler methods become more cost-effective.
\end{itemize}

Additional visualization of model-specific results can be found in Appendix Section \ref{Model-Specific}.

\begin{figure}[t]
    \centering
    \begin{tikzpicture}
    \begin{axis}[
        xlabel={Cost Concern ($C$)},
        ylabel={\textsc{EPI}},
        xtick=data,
        xticklabels={No Cost, Slight, Moderate, Elevated, Major},
        cycle list name=color list,
        grid=major,
        grid style=dashed,
        width=\columnwidth,
        height=6cm,
        tick pos=left,
        ymajorgrids=true,
        legend style={
            at={(0.5,-0.25)}, 
            anchor=north, 
            legend columns=2, 
            font=\small 
        },
        label style={font=\small},
        tick label style={font=\small},
        every axis plot/.append style={thick, mark=*} 
    ]

    \addplot[COT, mark=*, mark options={fill=COT}] coordinates {(0, 0.89) (1, 0.83460951) (2, 0.7826663305) (3, 0.6882770616) (4, 0.5322756332)};
    \addlegendentry{Chain-of-Thought}

    \addplot[SC, mark=*, mark options={fill=SC}] coordinates {(0, 0.95) (1, 0.7830562125) (2, 0.6454495073) (3, 0.4385316489) (4, 0.2024315864)};
    \addlegendentry{Self-Consistency}

    \addplot[TREE, mark=*, mark options={fill=TREE}] coordinates {(0, 0.79) (1, 0.7192240665) (2, 0.6547889339) (3, 0.542719681) (4, 0.3728413319)};
    \addlegendentry{Tree of Thoughts}

    \addplot[TOT, mark=*, mark options={fill=TOT}] coordinates {(0, 0.89) (1, 0.8157039337) (2, 0.7476100083) (3, 0.6280008141) (4, 0.4431292387)};
    \addlegendentry{Thread of Thought}

    \addplot[STANDARD, mark=*, mark options={fill=STANDARD}] coordinates {(0, 0.86) (1, 0.8143944975) (2, 0.771207439) (3, 0.6915824581) (4, 0.5561468562)};
    \addlegendentry{Standard}

    \addplot[S2A, mark=*, mark options={fill=S2A}] coordinates {(0, 0.68) (1, 0.6224434608) (2, 0.5697586204) (3, 0.4773895375) (4, 0.335148192)};
    \addlegendentry{System 2 Attention}

    \end{axis}
    \end{tikzpicture}
    \vspace{-1mm}
    \caption{EPI by prompt method on GSM8K, relative to cost concern \& averaged across all models}
    \label{fig:GSM8K-Line}
\end{figure}
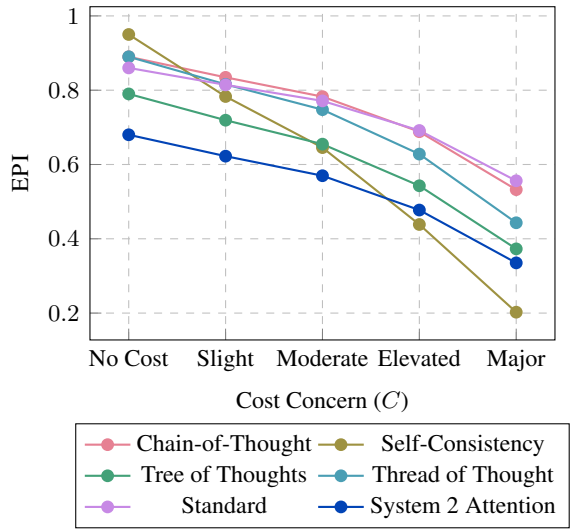



\section{Case Studies}
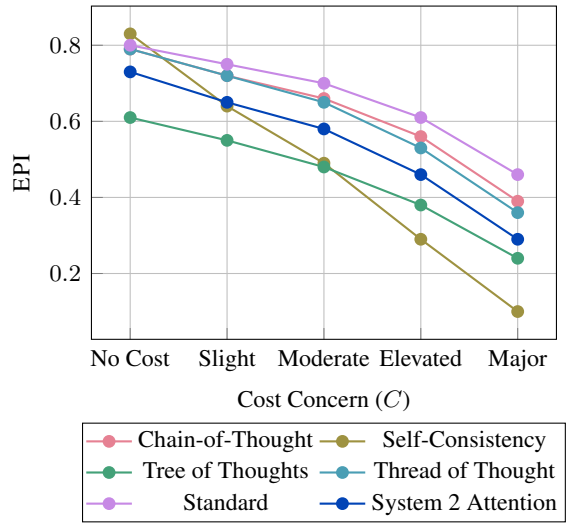
\begin{figure}[t]
    \centering
        \begin{tikzpicture}
        \begin{axis}[
            xlabel={Cost Concern ($C$)},
            ylabel={\textsc{EPI}},
            xtick=data,
            xticklabels={No Cost, Slight, Moderate, Elevated, Major},
            cycle list name=color list,
            grid=major,
            width=\columnwidth,
            height=6cm,
            tick pos=left,
            ymajorgrids=true,
            legend style={
                at={(0.5,-0.25)},
                anchor=north,
                legend columns=2,
                font=\small,
                inner sep=1pt,
                nodes={scale=1, transform shape}
            },
            label style={font=\small},
            tick label style={font=\small},
            every axis plot/.append style={thick, mark=*}
        ]

        \addplot[COT, mark=*, mark options={fill=COT}] coordinates {(0, 0.79) (1, 0.72) (2, 0.66) (3, 0.56) (4, 0.39)};
        \addlegendentry{Chain-of-Thought}

        \addplot[SC, mark=*, mark options={fill=SC}] coordinates {(0, 0.83) (1, 0.64) (2, 0.49) (3, 0.29) (4, 0.10)};
        \addlegendentry{Self-Consistency}

        \addplot[TREE, mark=*, mark options={fill=TREE}] coordinates {(0, 0.61) (1, 0.55) (2, 0.48) (3, 0.38) (4, 0.24)};
        \addlegendentry{Tree of Thoughts}

        \addplot[TOT, mark=*, mark options={fill=TOT}] coordinates {(0, 0.79) (1, 0.72) (2, 0.65) (3, 0.53) (4, 0.36)};
        \addlegendentry{Thread of Thought}

        \addplot[STANDARD, mark=*, mark options={fill=STANDARD}] coordinates {(0, 0.80) (1, 0.75) (2, 0.70) (3, 0.61) (4, 0.46)};
        \addlegendentry{Standard}

        \addplot[S2A, mark=*, mark options={fill=S2A}] coordinates {(0, 0.73) (1, 0.65) (2, 0.58) (3, 0.46) (4, 0.29)};
        \addlegendentry{System 2 Attention}

        \end{axis}
        \end{tikzpicture}
        \vspace{-1mm}
        \caption{EPI by prompt method on Claude 3.5 Sonnet, relative to cost concern \& averaged across all tasks}
        \label{fig:claude3.5-line}
\end{figure}
\label{Case Studies}

To demonstrate the practical utility of the EPI in real-world scenarios, we present two contrasting case studies that illustrate how organizations with different priorities and constraints can use the metric to make informed decisions about prompting strategies.

\subsection{Case Study 1: Optimizing Cost-Efficiency for a Large-Scale Virtual Assistant}
Company X, a leading provider of AI-powered customer service solutions, currently uses GPT-4 for their virtual assistant platform, serving over 500 enterprise clients at \$45/1M tokens.\footnote{https://openai.com/api/pricing/} The virtual assistant handles approximately 1 million customer inquiries daily.

\begin{figure}[t]
    \centering
        \begin{tikzpicture}
        \begin{axis}[
            xlabel={Cost Concern ($C$)},
            ylabel={\textsc{EPI}},
            xtick=data,
            xticklabels={No Cost, Slight, Moderate, Elevated, Major},
            cycle list name=color list,
            grid=major,
            width=\columnwidth,
            height=6cm,
            tick pos=left,
            ymajorgrids=true,
            legend style={
                at={(0.5,-0.25)},
                anchor=north,
                legend columns=2,
                font=\small,
                inner sep=1pt,
                nodes={scale=1, transform shape}
            },
            label style={font=\small},
            tick label style={font=\small},
            every axis plot/.append style={thick, mark=*}
        ]

        \addplot[COT, mark=*, mark options={fill=COT}] coordinates {(0, 0.89) (1, 0.83) (2, 0.78) (3, 0.69) (4, 0.53)};
        \addlegendentry{Chain-of-Thought}

        \addplot[STANDARD, mark=*, mark options={fill=STANDARD}] coordinates {(0, 0.86) (1, 0.83) (2, 0.80) (3, 0.75) (4, 0.65)};
        \addlegendentry{Standard}

        \end{axis}
        \end{tikzpicture}
        \vspace{-1mm}
        \caption{EPI comparison between Chain-of-Thought and standard prompting, given the parameters in Case Study 1.}
        \label{fig:casestudy1figure}
\end{figure}
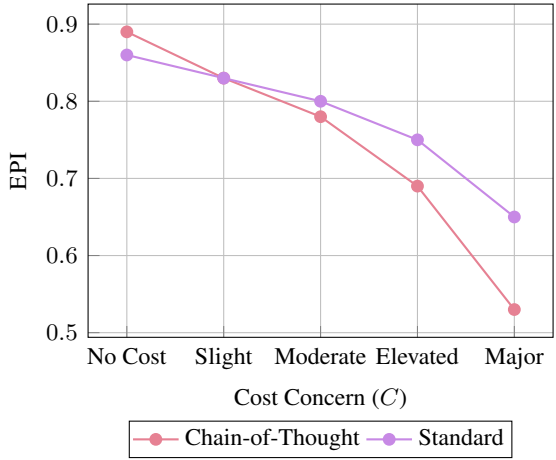

To optimize their system, Company X conducts an EPI analysis comparing their current Chain-of-Thought prompting (257 tokens/query, 0.89 accuracy) against standard prompting (137 tokens/query, 0.86 accuracy). As shown in Figure \ref{fig:casestudy1figure}, standard prompting demonstrates superior cost-efficiency, with the approaches intersecting at a very low cost concern ($C=0.00029$).

By switching to standard prompting, Company X projects:
\begin{itemize}
    \item A 47\% reduction in token consumption
    \item Annual savings of \$134,700 based on current usage
    \item Maintained performance levels (accuracy drop from 0.89 to 0.86)
\end{itemize}

\subsection{Case Study 2: Enhancing Performance for a Product Recommendation System}

Company Y, a mid-sized e-commerce platform, uses Claude 3.5 Haiku for their recommendation system at \$0.75/1M tokens.\footnote{https://www.anthropic.com/pricing\#anthropic-api} Their current standard prompting approach shows an average token consumption of 159 with 0.43 accuracy, while Chain-of-Thought prompting shows 242 tokens with 0.56 accuracy.

\begin{figure}[t]
    \centering
        \begin{tikzpicture}
        \begin{axis}[
            xlabel={Cost Concern ($C$)},
            ylabel={\textsc{EPI}},
            xtick=data,
            xticklabels={No Cost, Slight, Moderate, Elevated, Major},
            cycle list name=color list,
            grid=major,
            width=\columnwidth,
            height=6cm,
            tick pos=left,
            ymajorgrids=true,
            legend style={
                at={(0.5,-0.25)},
                anchor=north,
                legend columns=2,
                font=\small,
                inner sep=1pt,
                nodes={scale=1, transform shape}
            },
            label style={font=\small},
            tick label style={font=\small},
            every axis plot/.append style={thick, mark=*}
        ]

        \addplot[COT, mark=*, mark options={fill=COT}] coordinates {(0, 0.56) (1, 0.53) (2, 0.50) (3, 0.44) (4, 0.35)};
        \addlegendentry{Chain-of-Thought}

        \addplot[STANDARD, mark=*, mark options={fill=STANDARD}] coordinates {(0, 0.43) (1, 0.41) (2, 0.40) (3, 0.37) (4, 0.31)};
        \addlegendentry{Standard}

        \end{axis}
        \end{tikzpicture}
        \vspace{-1mm}
        \caption{EPI comparison between Chain-of-Thought and standard prompting, given the parameters in Case Study 2.}
        \label{fig:casestudy2figure}
\end{figure}
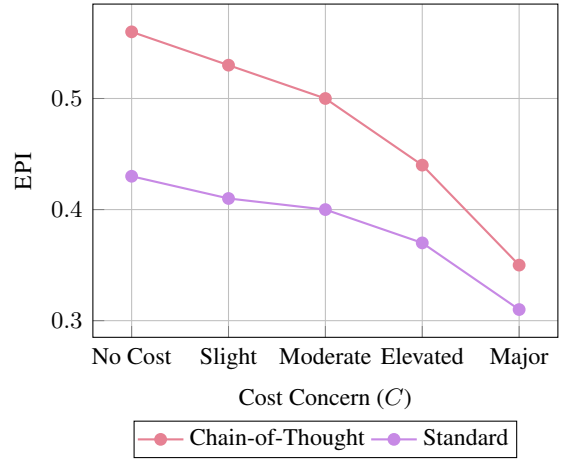

As illustrated in Figure \ref{fig:casestudy2figure}, the EPI analysis reveals that the benefits of standard prompting are only realized at high levels of cost concern (intersection at $C=0.00318$), while Chain-of-Thought's performance gains outweigh its cost implications for most practical purposes. By implementing Chain-of-Thought prompting, Company Y projects:
\begin{itemize}
    \item A 30\% increase in recommendation accuracy
    \item A manageable 52\% increase in token usage, justified by the performance gains
\end{itemize}

\section{Related Work}

Recent work has explored diverse prompting strategies to enhance LLM performance, including question decomposition, recursive reasoning, and programmatic decomposition \citep{wei2023chainofthought, wang2023selfconsistency, yao2023tree, weston20232, zhou2023thread, gao2023pal}. However, few studies have examined these techniques through the dual lens of performance and resource usage \citep{taherkhani2024epiccosteffectivesearchbasedprompt, nananukul2024costefficientpromptengineeringunsupervised, wang2024reasoningtokeneconomiesbudgetaware}. Surveys of the field have noted a disproportionate focus on accuracy as the primary metric in most studies \citep{sahoo2024systematic, vatsal2024surveypromptengineeringmethods}. While some efforts have been made to create more efficient versions of existing methods, such as Concise COT prompting \citep{renze2024benefits}, there remains a tendency to prioritize incremental performance gains without adequately considering the associated resource overhead. 

\section{Conclusion}

We introduce the Economical Prompting Index (EPI), a metric balancing accuracy and token usage in prompt evaluation. Our study across diverse datasets and models shows that while techniques like Self-Consistency often achieve higher accuracy, simpler methods can be more cost-effective under resource constraints. The EPI offers a flexible tool for assessing prompting techniques' practical viability in various scenarios. As LLMs evolve, metrics like the EPI will be crucial for developing accessible and efficient AI solutions. Future work could extend the EPI's application to broader tasks and models, and explore its role in creating resource-aware prompting techniques.

\section*{Limitations}

\paragraph{Temporal Validity of Results:} The field of LLMs and prompt design and optimization is rapidly evolving. Our results reflect the state of the art at the time of the study, but new models, prompting techniques, or optimization methods could emerge that significantly alter the landscape. This dynamic nature of the field means that the relative performance and efficiency of different techniques may change over time, potentially affecting the long-term applicability of our current findings.

\paragraph{Simplification of Cost Metrics:} The EPI uses token count as a proxy for computational cost. While this provides a straightforward and comparable metric across different models and techniques, it may not capture all aspects of real-world implementation costs. Factors such as inference time, memory usage, or model-specific pricing structures are not directly accounted for in our current formulation of the EPI. This simplification, while necessary for broad comparability, may not fully reflect cost considerations in all practical applications.

\paragraph{Generalizability Across Tasks:} Our study focuses on a specific set of task types represented by the chosen datasets. While these cover important areas such as mathematics, common sense reasoning, and multitask understanding, they may not encompass the full range of tasks for which LLMs are employed. The effectiveness of different prompting techniques, and consequently their EPI scores, may vary for more specialized or complex real-world applications not represented in our current task set.

\paragraph{The Hidden 
\textit{Cost} of Performance Reduction:} While the EPI considers computational efficiency, it doesn't account for potential financial impacts of reduced performance. In some applications, a small decrease in accuracy could have significant economic consequences (e.g., in financial forecasting or medical diagnosis) that might outweigh the computational cost savings.

\paragraph{Token Pricing Simplification:} Our current EPI implementation treats all tokens equally. However, in many LLM services, input tokens (prompts) are priced higher than output tokens (responses). This simplification in our model may not fully reflect the varied pricing structures in real-world LLM applications.

\paragraph{Limited Cost Concern Levels:} We provided five sample levels of cost concern in our analysis. However, this may not cover the full spectrum of real-world scenarios. Future work could explore a wider range of cost concern levels and incorporate user studies to better understand typical constraints in various applications.

\paragraph{Model and Machine Level Metrics:} The EPI is designed to evaluate the level of accuracy and cost of various prompts, but does not include considerations for advanced model metrics, such as inference time, or machine-level metrics, such as power consumption. Due to the closed-source nature of multiple models included for testing, machine-level metrics cannot feasibly be evaluated; future work could explore an additional study into response times with respect to various prompts.
\section*{Acknowledgments}
This work was supported by the Natural Sciences and Engineering Research Council of Canada and by the New Frontiers in Research Fund. Tyler McDonald and Anthony Colosimo were supported by the Natural Sciences and Engineering Research Council of Canada's Undergraduate Student Research Award. Yifeng Li is a Tier 2 Canada Research Chair in Machine Learning for Biomedical Data Science.

\bibliography{custom}

\clearpage
\appendix
\onecolumn
\section*{Appendix}
\section{Visualization of Linear and Polynomial EPI Approaches}
\label{sec:epi-appendix}

\begin{figure}[!htbp]
    \centering
    \begin{minipage}{0.48\textwidth} 
        \centering
        \begin{tikzpicture}
        \begin{axis}[
            xlabel={Token Count ($T$)},
            ylabel={\textsc{EPI} ($y$)},
            legend style={
                at={(0.5,-0.3)}, 
                anchor=north, 
                legend columns=1, 
                font=\small 
            },
            grid style=dashed,
            grid=both,
            width=\columnwidth, 
            height=6cm 
        ]

        \def\Y{1}

        \pgfmathdeclarefunction{Yadjusted}{3}{%
            \pgfmathparse{max(0, #1 - #3 * #2)}
        }

        \addplot [domain=0:1200, samples=100, thick, blue] {Yadjusted(\Y, x, 0.00083)};
        \addlegendentry{Major Cost Concern (C = 0.00083)}

        \addplot [domain=0:1200, samples=100, thick, orange] {Yadjusted(\Y, x, 0.0005)};
        \addlegendentry{Elevated Cost Concern (C = 0.0005)}

        \addplot [domain=0:1200, samples=100, thick, green] {Yadjusted(\Y, x, 0.00025)};
        \addlegendentry{Moderate Cost Concern (C = 0.00025)}

        \addplot [domain=0:1200, samples=100, thick, red] {Yadjusted(\Y, x, 0.0001)};
        \addlegendentry{Slight Cost Concern (C = 0.0001)}

        \addplot [domain=0:1200, samples=100, thick, purple] {Yadjusted(\Y, x, 0)};
        \addlegendentry{No Cost Concern (C = 0)}
        \end{axis}
        \end{tikzpicture}
        \caption{Graph of input token count $T$ against a linear EPI calculation ($EPI = \max(0, A - C \cdot T)$), given the weight classes $C$ for $A=1$.}
        \label{fig:linear-plot}
    \end{minipage}%
    \hfill
    \vspace{1cm}
    \begin{minipage}{0.48\textwidth} 
        \centering
        \begin{tikzpicture}
        \begin{axis}[
            xlabel={Token Count ($T$)},
            ylabel={\textsc{EPI} ($y$)},
            legend style={
                at={(0.5,-0.3)}, 
                anchor=north, 
                legend columns=1, 
                font=\small 
            },
            grid style=dashed,
            grid=both,
            width=\columnwidth, 
            height=6cm 
        ]

        \def\Y{1}

        \pgfmathdeclarefunction{Yadjusted}{3}{%
            \pgfmathparse{max(0, #1 - #3 * #2^2)}
        }

        \addplot [domain=0:1200, samples=100, thick, blue] {Yadjusted(\Y, x, 1.5e-6)};
        \addlegendentry{Major Cost Concern (C = $1.5 \times 10^{-6}$)}

        \addplot [domain=0:1200, samples=100, thick, orange] {Yadjusted(\Y, x, 7e-7)};
        \addlegendentry{Elevated Cost Concern (C = $7 \times 10^{-7}$)}

        \addplot [domain=0:1200, samples=100, thick, green] {Yadjusted(\Y, x, 4e-7)};
        \addlegendentry{Moderate Cost Concern (C = $4 \times 10^{-7}$)}

        \addplot [domain=0:1200, samples=100, thick, red] {Yadjusted(\Y, x, 2e-7)};
        \addlegendentry{Slight Cost Concern (C = $2 \times 10^{-7}$)}

        \addplot [domain=0:1200, samples=100, thick, purple] {Yadjusted(\Y, x, 0)};
        \addlegendentry{No Cost Concern (C = $0$)}
        \end{axis}
        \end{tikzpicture}
        \caption{Graph of input token count $T$ against a quadratic EPI calculation ($EPI = \max(0, A - C \cdot T^2)$), given the weight classes $C$ for $A=1$.}
        \label{fig:quadratic-plot}
    \end{minipage}
\end{figure}

\section{Prompt Input Texts}
\label{Prompt Text}
\textbf{Chain-of-Thought}: "<question>. Let's think step-by-step."\\
\textbf{Self-Consistency}: \textit{Chain-of-Thought sampled 3 times, highest occurring result marginalized.}\\
\textbf{Tree of Thoughts}: "Imagine three different experts are answering this question.
All experts will write down 1 step of their thinking,
then share it with the group.
Then all experts will go on to the next step, etc.
If any expert realizes they're wrong at any point, then they leave.
The question is <question>."\\
\textbf{System 2 Attention}: "Given the following text by a user, extract the part that is unbiased and not their opinion,
so that using that text alone would be good context for providing an unbiased answer to
the question portion of the text.
Please include the actual question or query that the user is asking. Separate this
into two categories labeled with “Unbiased text context (includes all content except user’s
bias):” and “Question/Query (does not include user bias/preference):After such, use this new unbiased text to answer the proposed question”. Text by User: {question}')”.
Text by User: <question>"\\
\textbf{Thread of Thought}: "Walk me through this context in manageable parts step by step, summarizing and analyzing as we go.
<question>"

\noindent For each multiple choice dataset (CSQA, DQA and MMLU) an additional statement was added to ensure that the response declared a final answer: "End your answer in this exact format: Final Answer = (LETTER) ex. Final Answer (B). The letter you are selecting for your final answer must be surrounded by 2 parentheses, ex. (A). Only do this once."

\noindent For the MMLU dataset, the declared answer had to be surrounded by 2 parentheses on each side. (ex. Final Answer = ((LETTER)).)

\newpage
\section{Example Outputs - CSQA}
\label{Examples}
\vspace{-5mm}
\textbf{Input}: The sanctions against the school were a punishing blow, and they seemed to what the efforts the school had made to change? A) ignore, B) enforce, C) authoritarian, D) yell at, E) avoid.\\\textbf{Standard}: Final Answer = (A)\\
\textbf{Chain-of-Thought}: Firstly, the sentence suggests that the sanctions have a negative impact on the school's efforts. Therefore, any word suggesting a positive impact, such as "enforce" (B), cannot be the right one. Secondly, the sentence is about the sanctions' impact, so a verb should be used here, meaning "authoritarian (C)" and "yell at (D)", which are an adjective and a phrase respectively, are not correct either. Thirdly, this sentence seems to suggest that the sanctions are not recognizing or taking into account the school's efforts at change. This leaves "ignore (A)" and "avoid (E)". Out of these two, "avoid" implies a deliberate action of staying away from the school's efforts, which is not really what the sentence is suggesting. So, "ignore (A)", which suggests the sanctions are not acknowledging the school's efforts at change, is the most suitable choice. Final Answer = (A)\\
\textbf{Self-Consistency}:\\
\textit{Answer 1}: The sentence is saying that the sanctions against the school were harsh and seemed to overlook or disregard the efforts the school had made to change. So, we want an option that can fit this theme. The options B), C), D) and E) are not fitting in this context. The best fit in this context is A) ignore. This portrays a situation where the school has put in an effort to make changes, but the sanctions were a blow to this effort, as if those efforts were ignored or not recognized. Therefore, the correct answer is A) ignore. Final Answer = (A)\\
\textit{Answer 2}: Firstly, we can eliminate C and D, as 'authoritarian' is not a verb and 'yell at' doesn't fit in this context. Secondly, 'enforce' and 'avoid' also doesn't make sense in this context. So, the remaining option is 'ignore'. The sanctions seem to disregard or ignore the efforts the school had made to change. Final Answer = (A)\\
\textit{Answer 3}: Firstly, we need a verb that suits the context. So, we rule out 'C' which is an adjective and 'D' which seems very informal. Secondly, we consider the meaning of the remaining options. 'Enforce' suggests that the sanction would strengthen the school's efforts, which doesn't align with the previous part of the sentence mentioning a 'punishing blow'. 'Avoid' would suggest the sanction were designed to dodge the efforts - again, not matching the context. 'Ignore' fits best, signifying that despite the school's efforts, the sanctions acted as though these efforts didn't matter. Final Answer = (A)\\
\textbf{Tree of Thoughts}: Expert 1: The question is asking us to find a word that fits the context of the sentence. The sanctions had an impact which didn't consider the school's efforts to change. Drawing from this, my initial thought is that the correct option will be a word meaning to disregard or not take into account. Final Answer = (A)\\Expert 2: Considering the thematic of the question, it's clear we are looking for a word that could show a lack of regard for the changes done by the school. The word that fits this description better from the options given is "ignore". Final Answer = (A)\\Expert 3: The sanctions are delivered to the school despite the actions it has taken to amend its ways, suggesting that the sanctions don't take into account these efforts. The word that matches with this concept from the options provided is "ignore". Final Answer = (A)\\
\textbf{System 2 Attention}: Unbiased text context (includes all content except user's bias): The sanctions against the school were a punishing blow. Question/Query (does not include user bias/preference): What did the sanctions against the school seem to do to the efforts the school had made to change?\\
\textbf{Thread of Thought}: The context describes a situation where the school is facing sanctions, described as a punishing blow. These sanctions appear to be ignoring or dismissing the efforts the school has already put into making changes. Now let's break down the possible answers one by one: A) Ignore: This answer could be correct as it fits the context. The sanctions might be neglecting the efforts made by the school. B) Enforce: this doesn't fit because sanctions enforce a penalty, not the efforts made by the school. C) Authoritarian: This isn't fitting as it's an adjective and doesn't make grammatical sense in the sentence. D) Yell at: This isn't plausible because it's not feasible to yell at the efforts themselves. E) Avoid: This is less likely as sanctions are not capable of avoiding anything. In summary, looking at the context and analyzing the answers, the most plausible answer is (A) Ignore. Final Answer = (A)

\section{Tests for Statistical Significance}
\label{Stat-Sig}
To evaluate the statistical significance of accuracy for a given prompting method, we conduct a Two Proportion Z-Test between the two highest-performing methods on each dataset. Similarly, to assess the statistical significance of cost, we use a Paired t-Test between the two most expensive methods on each dataset. For each dataset, significance tests are performed on the samples from each of the ten models; a result is reported as significant only if it is significant on all models. All tests are conducted at a 95\% confidence interval ($p<0.05$).

\section{Model-Agnostic Figures}
\label{Model-Agnostic}

\begin{figure}[!htbp]
    \centering
    \begin{minipage}{0.48\textwidth}
        \centering
        \begin{tikzpicture}
        \begin{axis}[
            xtick=data,
            xticklabels={No Cost, Slight, Moderate, Elevated, Major},
            cycle list name=color list,
            grid=major,
            width=\columnwidth,
            height=6cm,
            tick pos=left,
            ymajorgrids=true,
            legend style={
                at={(0.5,-0.2)},
                anchor=north,
                legend columns=2,
                font=\small
            },
            label style={font=\small},
            tick label style={font=\small},
            every axis plot/.append style={thick, mark=*}
        ]

        \addplot[COT, mark=*, mark options={fill=COT}] coordinates {(0, 0.739) (1, 0.686) (2, 0.636) (3, 0.547) (4, 0.404)};
        \addlegendentry{Chain-of-Thought}

        \addplot[SC, mark=*, mark options={fill=SC}] coordinates {(0, 0.836) (1, 0.667) (2, 0.532) (3, 0.339) (4, 0.137)};
        \addlegendentry{Self-Consistency}

        \addplot[TREE, mark=*, mark options={fill=TREE}] coordinates {(0, 0.660) (1, 0.593) (2, 0.533) (3, 0.431) (4, 0.281)};
        \addlegendentry{Tree of Thoughts}

        \addplot[TOT, mark=*, mark options={fill=TOT}] coordinates {(0, 0.731) (1, 0.658) (2, 0.593) (3, 0.481) (4, 0.317)};
        \addlegendentry{Thread of Thought}

        \addplot[STANDARD, mark=*, mark options={fill=STANDARD}] coordinates {(0, 0.753) (1, 0.713) (2, 0.674) (3, 0.604) (4, 0.484)};
        \addlegendentry{Standard}

        \addplot[S2A, mark=*, mark options={fill=S2A}] coordinates {(0, 0.625) (1, 0.565) (2, 0.511) (3, 0.418) (4, 0.280)};
        \addlegendentry{System 2 Attention}

        \end{axis}
        \end{tikzpicture}
        \caption{MMLU \textsc{Economical Prompt Indexes} for each prompt method by cost concern taxonomy.}
        \label{fig:MMLU-Line}
    \end{minipage}
    \hfill
    \begin{minipage}{0.48\textwidth}
        \centering
        \begin{tikzpicture}
        \begin{axis}[
            xtick=data,
            xticklabels={No Cost, Slight, Moderate, Elevated, Major},
            cycle list name=color list,
            grid=major,
            width=\columnwidth,
            height=6cm,
            tick pos=left,
            ymajorgrids=true,
            legend style={
                at={(0.5,-0.2)},
                anchor=north,
                legend columns=2,
                font=\small
            },
            label style={font=\small},
            tick label style={font=\small},
            every axis plot/.append style={thick, mark=*}
        ]

        \addplot[COT, mark=*, mark options={fill=COT}] coordinates {(0, 0.603) (1, 0.569) (2, 0.537) (3, 0.479) (4, 0.381)};
        \addlegendentry{Chain-of-Thought}

        \addplot[SC, mark=*, mark options={fill=SC}] coordinates {(0, 0.757) (1, 0.637) (2, 0.536) (3, 0.380) (4, 0.190)};
        \addlegendentry{Self-Consistency}

        \addplot[TREE, mark=*, mark options={fill=TREE}] coordinates {(0, 0.603) (1, 0.548) (2, 0.497) (3, 0.410) (4, 0.279)};
        \addlegendentry{Tree of Thoughts}

        \addplot[TOT, mark=*, mark options={fill=TOT}] coordinates {(0, 0.602) (1, 0.562) (2, 0.525) (3, 0.458) (4, 0.348)};
        \addlegendentry{Thread of Thought}

        \addplot[STANDARD, mark=*, mark options={fill=STANDARD}] coordinates {(0, 0.578) (1, 0.555) (2, 0.533) (3, 0.492) (4, 0.418)};
        \addlegendentry{Standard}

        \addplot[S2A, mark=*, mark options={fill=S2A}] coordinates {(0, 0.448) (1, 0.409) (2, 0.373) (3, 0.311) (4, 0.216)};
        \addlegendentry{System 2 Attention}

        \end{axis}
        \end{tikzpicture}
        \caption{Disambiguation QA \textsc{Economical Prompt Indexes} for each prompt method.}
        \label{fig:DQA-Line}
    \end{minipage}

    \vspace{1cm} 
    \begin{minipage}{0.48\textwidth}
        \centering
        \begin{tikzpicture}
        \begin{axis}[
            xtick=data,
            xticklabels={No Cost, Slight, Moderate, Elevated, Major},
            cycle list name=color list,
            grid=major,
            width=\columnwidth,
            height=6cm,
            tick pos=left,
            ymajorgrids=true,
            legend style={
                at={(0.5,-0.2)},
                anchor=north,
                legend columns=2,
                font=\small
            },
            label style={font=\small},
            tick label style={font=\small},
            every axis plot/.append style={thick, mark=*}
        ]

        \addplot[COT, mark=*, mark options={fill=COT}] coordinates {(0, 0.790) (1, 0.750) (2, 0.713) (3, 0.643) (4, 0.524)};
        \addlegendentry{Chain-of-Thought}

        \addplot[SC, mark=*, mark options={fill=SC}] coordinates {(0, 0.883) (1, 0.756) (2, 0.647) (3, 0.475) (4, 0.255)};
        \addlegendentry{Self-Consistency}

        \addplot[TREE, mark=*, mark options={fill=TREE}] coordinates {(0, 0.741) (1, 0.673) (2, 0.612) (3, 0.505) (4, 0.344)};
        \addlegendentry{Tree of Thoughts}

        \addplot[TOT, mark=*, mark options={fill=TOT}] coordinates {(0, 0.782) (1, 0.721) (2, 0.665) (3, 0.565) (4, 0.409)};
        \addlegendentry{Thread of Thought}

        \addplot[STANDARD, mark=*, mark options={fill=STANDARD}] coordinates {(0, 0.772) (1, 0.745) (2, 0.719) (3, 0.670) (4, 0.582)};
        \addlegendentry{Standard}

        \addplot[S2A, mark=*, mark options={fill=S2A}] coordinates {(0, 0.673) (1, 0.624) (2, 0.578) (3, 0.497) (4, 0.367)};
        \addlegendentry{System 2 Attention}

        \end{axis}
        \end{tikzpicture}
        \caption{Common Sense QA \textsc{Economical Prompt Indexes} for each prompt method.}
        \label{fig:csqa-line}
    \end{minipage}
\end{figure}

\newpage

\section{Model-Specific Figures}
\label{Model-Specific}

\begin{figure}[!htbp]
    \centering
    \begin{minipage}{0.48\textwidth}
        \centering
        \begin{tikzpicture}
        \begin{axis}[
            xtick=data,
            xticklabels={No Cost, Slight, Moderate, Elevated, Major},
            cycle list name=color list,
            grid=major,
            width=\columnwidth,
            height=6cm,
            tick pos=left,
            ymajorgrids=true,
            legend style={
                at={(0.5,-0.2)},
                anchor=north,
                legend columns=2,
                font=\small,
                inner sep=1pt,
                nodes={scale=1, transform shape}
            },
            label style={font=\small},
            tick label style={font=\small},
            every axis plot/.append style={thick, mark=*}
        ]

        \addplot[COT, mark=*, mark options={fill=COT}] coordinates {(0, 0.73) (1, 0.69) (2, 0.66) (3, 0.59) (4, 0.48)};
        \addlegendentry{Chain-of-Thought}

        \addplot[SC, mark=*, mark options={fill=SC}] coordinates {(0, 0.88) (1, 0.75) (2, 0.64) (3, 0.47) (4, 0.25)};
        \addlegendentry{Self-Consistency}

        \addplot[TREE, mark=*, mark options={fill=TREE}] coordinates {(0, 0.73) (1, 0.68) (2, 0.63) (3, 0.55) (4, 0.41)};
        \addlegendentry{Tree of Thoughts}

        \addplot[TOT, mark=*, mark options={fill=TOT}] coordinates {(0, 0.66) (1, 0.62) (2, 0.58) (3, 0.51) (4, 0.40)};
        \addlegendentry{Thread of Thought}

        \addplot[STANDARD, mark=*, mark options={fill=STANDARD}] coordinates {(0, 0.70) (1, 0.67) (2, 0.65) (3, 0.61) (4, 0.53)};
        \addlegendentry{Standard}

        \addplot[S2A, mark=*, mark options={fill=S2A}] coordinates {(0, 0.64) (1, 0.59) (2, 0.54) (3, 0.46) (4, 0.34)};
        \addlegendentry{System 2 Attention}

        \end{axis}
        \end{tikzpicture}
        \caption{\textsc{Economical Prompt Indexes} for each prompt method tested on GPT-3.5-Turbo.}
        \label{fig:gpt35-line}
    \end{minipage}
    \hfill
    \begin{minipage}{0.48\textwidth}
    \centering
    \begin{tikzpicture}
    \begin{axis}[
        xlabel={Cost Concern ($C$)},
        ylabel={\textsc{EPI} ($y$)},
        xtick=data,
        xticklabels={No Cost, Slight, Moderate, Elevated, Major},
        cycle list name=color list,
        grid=major,
        grid style=dashed,
        width=\columnwidth,
        height=6cm,
        tick pos=left,
        ymajorgrids=true,
        legend style={
            at={(0.5,-0.275)},
            anchor=north,
            legend columns=2,
            font=\small,
            inner sep=1pt,
            nodes={scale=1, transform shape}
        },
        label style={font=\small},
        tick label style={font=\small},
        every axis plot/.append style={thick, mark=*}
    ]
    \addplot[COT, mark=*, mark options={fill=COT}] coordinates {(0, 0.83) (1, 0.79) (2, 0.75) (3, 0.67) (4, 0.54)};
    \addlegendentry{Chain-of-Thought}
    \addplot[SC, mark=*, mark options={fill=SC}] coordinates {(0, 0.90) (1, 0.76) (2, 0.65) (3, 0.46) (4, 0.24)};
    \addlegendentry{Self-Consistency}
    \addplot[TREE, mark=*, mark options={fill=TREE}] coordinates {(0, 0.84) (1, 0.77) (2, 0.70) (3, 0.59) (4, 0.42)};
    \addlegendentry{Tree of Thoughts}
    \addplot[TOT, mark=*, mark options={fill=TOT}] coordinates {(0, 0.85) (1, 0.78) (2, 0.72) (3, 0.61) (4, 0.44)};
    \addlegendentry{Thread of Thought}
    \addplot[STANDARD, mark=*, mark options={fill=STANDARD}] coordinates {(0, 0.84) (1, 0.81) (2, 0.79) (3, 0.74) (4, 0.64)};
    \addlegendentry{Standard}
    \addplot[S2A, mark=*, mark options={fill=S2A}] coordinates {(0, 0.29) (1, 0.27) (2, 0.25) (3, 0.22) (4, 0.16)};
    \addlegendentry{System 2 Attention}
    \end{axis}
    \end{tikzpicture}
    \vspace{-1mm}
    \caption{EPI across different prompting methods tested on GPT-4.}
    \label{fig:gpt4-line}
    \end{minipage}
\end{figure}

\begin{figure}[!htbp]
    \centering
    \begin{minipage}{0.48\textwidth}
        \centering
        \begin{tikzpicture}
        \begin{axis}[
            xtick=data,
            xticklabels={No Cost, Slight, Moderate, Elevated, Major},
            cycle list name=color list,
            grid=major,
            width=\columnwidth,
            height=6cm,
            tick pos=left,
            ymajorgrids=true,
            legend style={
                at={(0.5,-0.2)},
                anchor=north,
                legend columns=2,
                font=\small,
                inner sep=1pt,
                nodes={scale=1, transform shape}
            },
            label style={font=\small},
            tick label style={font=\small},
            every axis plot/.append style={thick, mark=*}
        ]

        \addplot[COT, mark=*, mark options={fill=COT}] coordinates {(0, 0.68) (1, 0.63) (2, 0.58) (3, 0.50) (4, 0.36)};
        \addlegendentry{Chain-of-Thought}

        \addplot[SC, mark=*, mark options={fill=SC}] coordinates {(0, 0.84) (1, 0.66) (2, 0.52) (3, 0.32) (4, 0.12)};
        \addlegendentry{Self-Consistency}

        \addplot[TREE, mark=*, mark options={fill=TREE}] coordinates {(0, 0.69) (1, 0.61) (2, 0.54) (3, 0.42) (4, 0.25)};
        \addlegendentry{Tree of Thoughts}

        \addplot[TOT, mark=*, mark options={fill=TOT}] coordinates {(0, 0.72) (1, 0.65) (2, 0.58) (3, 0.48) (4, 0.32)};
        \addlegendentry{Thread of Thought}

        \addplot[STANDARD, mark=*, mark options={fill=STANDARD}] coordinates {(0, 0.68) (1, 0.63) (2, 0.59) (3, 0.52) (4, 0.39)};
        \addlegendentry{Standard}

        \addplot[S2A, mark=*, mark options={fill=S2A}] coordinates {(0, 0.46) (1, 0.42) (2, 0.38) (3, 0.31) (4, 0.21)};
        \addlegendentry{System 2 Attention}

        \end{axis}
        \end{tikzpicture}
        \caption{\textsc{Economical Prompt Indexes} for each prompt method tested on Mixtral 8-7B.}
        \label{fig:mixtral7b-line}
    \end{minipage}
    \hfill
    \begin{minipage}{0.48\textwidth}
        \centering
        \begin{tikzpicture}
        \begin{axis}[
            xtick=data,
            xticklabels={No Cost, Slight, Moderate, Elevated, Major},
            cycle list name=color list,
            grid=major,
            width=\columnwidth,
            height=6cm,
            tick pos=left,
            ymajorgrids=true,
            legend style={
                at={(0.5,-0.2)},
                anchor=north,
                legend columns=2,
                font=\small,
                inner sep=1pt,
                nodes={scale=1, transform shape}
            },
            label style={font=\small},
            tick label style={font=\small},
            every axis plot/.append style={thick, mark=*}
        ]

        \addplot[COT, mark=*, mark options={fill=COT}] coordinates {(0, 0.79) (1, 0.73) (2, 0.68) (3, 0.59) (4, 0.44)};
        \addlegendentry{Chain-of-Thought}

        \addplot[SC, mark=*, mark options={fill=SC}] coordinates {(0, 0.89) (1, 0.71) (2, 0.57) (3, 0.36) (4, 0.15)};
        \addlegendentry{Self-Consistency}

        \addplot[TREE, mark=*, mark options={fill=TREE}] coordinates {(0, 0.86) (1, 0.76) (2, 0.66) (3, 0.51) (4, 0.30)};
        \addlegendentry{Tree of Thoughts}

        \addplot[TOT, mark=*, mark options={fill=TOT}] coordinates {(0, 0.83) (1, 0.74) (2, 0.67) (3, 0.53) (4, 0.34)};
        \addlegendentry{Thread of Thought}

        \addplot[STANDARD, mark=*, mark options={fill=STANDARD}] coordinates {(0, 0.79) (1, 0.74) (2, 0.69) (3, 0.60) (4, 0.45)};
        \addlegendentry{Standard}

        \addplot[S2A, mark=*, mark options={fill=S2A}] coordinates {(0, 0.74) (1, 0.66) (2, 0.58) (3, 0.45) (4, 0.28)};
        \addlegendentry{System 2 Attention}

        \end{axis}
        \end{tikzpicture}
        \caption{\textsc{Economical Prompt Indexes} for each prompt method tested on Mixtral 8-22B.}
        \label{fig:mixtral22b-line}

    \end{minipage}
\end{figure}

\begin{figure}[!htbp]
    \centering
    \begin{minipage}{0.48\textwidth}
    \centering
    \begin{tikzpicture}
    \begin{axis}[
        xtick=data,
        xticklabels={No Cost, Slight, Moderate, Elevated, Major},
        cycle list name=color list,
        grid=major,
        width=\columnwidth,
        height=6cm,
        tick pos=left,
        ymajorgrids=true,
        legend style={
            at={(0.5,-0.2)},
            anchor=north,
            legend columns=2,
            font=\small,
            inner sep=1pt,
            nodes={scale=1, transform shape}
        },
        label style={font=\small},
        tick label style={font=\small},
        every axis plot/.append style={thick, mark=*}
    ]

    \addplot[COT, mark=*, mark options={fill=COT}] coordinates {(0, 0.74) (1, 0.69) (2, 0.64) (3, 0.56) (4, 0.42)};
    \addlegendentry{Chain-of-Thought}

    \addplot[SC, mark=*, mark options={fill=SC}] coordinates {(0, 0.84) (1, 0.68) (2, 0.55) (3, 0.36) (4, 0.15)};
    \addlegendentry{Self-Consistency}

    \addplot[TREE, mark=*, mark options={fill=TREE}] coordinates {(0, 0.67) (1, 0.61) (2, 0.56) (3, 0.48) (4, 0.34)};
    \addlegendentry{Tree of Thoughts}

    \addplot[TOT, mark=*, mark options={fill=TOT}] coordinates {(0, 0.73) (1, 0.66) (2, 0.60) (3, 0.50) (4, 0.35)};
    \addlegendentry{Thread of Thought}

    \addplot[STANDARD, mark=*, mark options={fill=STANDARD}] coordinates {(0, 0.72) (1, 0.68) (2, 0.64) (3, 0.58) (4, 0.46)};
    \addlegendentry{Standard}

    \addplot[S2A, mark=*, mark options={fill=S2A}] coordinates {(0, 0.66) (1, 0.60) (2, 0.54) (3, 0.45) (4, 0.30)};
    \addlegendentry{System 2 Attention}

    \end{axis}
    \end{tikzpicture}
    \caption{\textsc{Economical Prompt Indexes} for each prompt method tested on Claude 3 Haiku.}
    \label{fig:claude3h-line}
    \end{minipage}
    \hfill
    \begin{minipage}{0.48\textwidth}
        \centering
        \begin{tikzpicture}
        \begin{axis}[
            xtick=data,
            xticklabels={No Cost, Slight, Moderate, Elevated, Major},
            cycle list name=color list,
            grid=major,
            width=\columnwidth,
            height=6cm,
            tick pos=left,
            ymajorgrids=true,
            legend style={
                at={(0.5,-0.2)},
                anchor=north,
                legend columns=2,
                font=\small,
                inner sep=1pt,
                nodes={scale=1, transform shape}
            },
            label style={font=\small},
            tick label style={font=\small},
            every axis plot/.append style={thick, mark=*}
        ]

        \addplot[COT, mark=*, mark options={fill=COT}] coordinates {(0, 0.80) (1, 0.75) (2, 0.71) (3, 0.63) (4, 0.49)};
        \addlegendentry{Chain-of-Thought}
        
        \addplot[SC, mark=*, mark options={fill=SC}] coordinates {(0, 0.88) (1, 0.74) (2, 0.61) (3, 0.43) (4, 0.21)};
        \addlegendentry{Self-Consistency}
        
        \addplot[TREE, mark=*, mark options={fill=TREE}] coordinates {(0, 0.73) (1, 0.66) (2, 0.61) (3, 0.51) (4, 0.35)};
        \addlegendentry{Tree of Thoughts}
        
        \addplot[TOT, mark=*, mark options={fill=TOT}] coordinates {(0, 0.78) (1, 0.72) (2, 0.66) (3, 0.56) (4, 0.40)};
        \addlegendentry{Thread of Thought}
        
        \addplot[STANDARD, mark=*, mark options={fill=STANDARD}] coordinates {(0, 0.78) (1, 0.74) (2, 0.70) (3, 0.63) (4, 0.51)};
        \addlegendentry{Standard}
        
        \addplot[S2A, mark=*, mark options={fill=S2A}] coordinates {(0, 0.70) (1, 0.64) (2, 0.59) (3, 0.49) (4, 0.33)};
        \addlegendentry{System 2 Attention}

        \end{axis}
        \end{tikzpicture}
        \caption{\textsc{Economical Prompt Indexes} for each prompt method tested on Gemini 1.5 Pro.}
        \label{fig:gemini1.5-line}
    \end{minipage}
\end{figure}

\begin{figure}[!htbp]
    \centering
    \begin{minipage}{0.48\textwidth}
        \centering
        \begin{tikzpicture}
        \begin{axis}[
            xtick=data,
            xticklabels={No Cost, Slight, Moderate, Elevated, Major},
            cycle list name=color list,
            grid=major,
            width=\columnwidth,
            height=6cm,
            tick pos=left,
            ymajorgrids=true,
            legend style={
                at={(0.5,-0.2)},
                anchor=north,
                legend columns=2,
                font=\small,
                inner sep=1pt,
                nodes={scale=1, transform shape}
            },
            label style={font=\small},
            tick label style={font=\small},
            every axis plot/.append style={thick, mark=*}
        ]

        \addplot[COT, mark=*, mark options={fill=COT}] coordinates {(0, 0.71) (1, 0.68) (2, 0.65) (3, 0.59) (4, 0.50)};
        \addlegendentry{Chain-of-Thought}

        \addplot[SC, mark=*, mark options={fill=SC}] coordinates {(0, 0.82) (1, 0.72) (2, 0.64) (3, 0.49) (4, 0.30)};
        \addlegendentry{Self-Consistency}

        \addplot[TREE, mark=*, mark options={fill=TREE}] coordinates {(0, 0.56) (1, 0.52) (2, 0.49) (3, 0.43) (4, 0.33)};
        \addlegendentry{Tree of Thoughts}

        \addplot[TOT, mark=*, mark options={fill=TOT}] coordinates {(0, 0.70) (1, 0.65) (2, 0.61) (3, 0.52) (4, 0.39)};
        \addlegendentry{Thread of Thought}

        \addplot[STANDARD, mark=*, mark options={fill=STANDARD}] coordinates {(0, 0.65) (1, 0.63) (2, 0.61) (3, 0.57) (4, 0.51)};
        \addlegendentry{Standard}

        \addplot[S2A, mark=*, mark options={fill=S2A}] coordinates {(0, 0.46) (1, 0.43) (2, 0.40) (3, 0.34) (4, 0.25)};
        \addlegendentry{System 2 Attention}

        \end{axis}
        \end{tikzpicture}
        \caption{\textsc{Economical Prompt Indexes} for each prompt method tested on Gemini 1.0 Pro.}
        \label{fig:gemini1-line}
    \end{minipage}
    \hfill
    \begin{minipage}{0.48\textwidth}
        \centering
        \begin{tikzpicture}
        \begin{axis}[
            xtick=data,
            xticklabels={No Cost, Slight, Moderate, Elevated, Major},
            cycle list name=color list,
            grid=major,
            width=\columnwidth,
            height=6cm,
            tick pos=left,
            ymajorgrids=true,
            legend style={
                at={(0.5,-0.2)},
                anchor=north,
                legend columns=2,
                font=\small,
                inner sep=1pt,
                nodes={scale=1, transform shape}
            },
            label style={font=\small},
            tick label style={font=\small},
            every axis plot/.append style={thick, mark=*}
        ]

        \addplot[COT, mark=*, mark options={fill=COT}] coordinates {(0, 0.82) (1, 0.76) (2, 0.71) (3, 0.62) (4, 0.47)};
        \addlegendentry{Chain-of-Thought}

        \addplot[SC, mark=*, mark options={fill=SC}] coordinates {(0, 0.90) (1, 0.73) (2, 0.60) (3, 0.39) (4, 0.17)};
        \addlegendentry{Self-Consistency}

        \addplot[TREE, mark=*, mark options={fill=TREE}] coordinates {(0, 0.65) (1, 0.55) (2, 0.48) (3, 0.35) (4, 0.19)};
        \addlegendentry{Tree of Thoughts}

        \addplot[TOT, mark=*, mark options={fill=TOT}] coordinates {(0, 0.79) (1, 0.71) (2, 0.64) (3, 0.52) (4, 0.34)};
        \addlegendentry{Thread of Thought}

        \addplot[STANDARD, mark=*, mark options={fill=STANDARD}] coordinates {(0, 0.77) (1, 0.74) (2, 0.72) (3, 0.67) (4, 0.58)};
        \addlegendentry{Standard}

        \addplot[S2A, mark=*, mark options={fill=S2A}] coordinates {(0, 0.75) (1, 0.69) (2, 0.62) (3, 0.52) (4, 0.35)};
        \addlegendentry{System 2 Attention}

        \end{axis}
        \end{tikzpicture}
        \caption{\textsc{Economical Prompt Indexes} for each prompt method tested on Llama 3-70B.}
        \label{fig:llama3-70b-line}
    \end{minipage}
\end{figure}

\begin{figure}[!htbp]
    \centering
    \begin{minipage}{0.48\textwidth}
        \centering
        \begin{tikzpicture}
        \begin{axis}[
            xtick=data,
            xticklabels={No Cost, Slight, Moderate, Elevated, Major},
            cycle list name=color list,
            grid=major,
            width=\columnwidth,
            height=6cm,
            tick pos=left,
            ymajorgrids=true,
            legend style={
                at={(0.5,-0.2)},
                anchor=north,
                legend columns=2,
                font=\small,
                inner sep=1pt,
                nodes={scale=1, transform shape}
            },
            label style={font=\small},
            tick label style={font=\small},
            every axis plot/.append style={thick, mark=*}
        ]

        \addplot[COT, mark=*, mark options={fill=COT}] coordinates {(0, 0.68) (1, 0.63) (2, 0.59) (3, 0.52) (4, 0.39)};
        \addlegendentry{Chain-of-Thought}

        \addplot[SC, mark=*, mark options={fill=SC}] coordinates {(0, 0.86) (1, 0.70) (2, 0.57) (3, 0.38) (4, 0.17)};
        \addlegendentry{Self-Consistency}

        \addplot[TREE, mark=*, mark options={fill=TREE}] coordinates {(0, 0.55) (1, 0.48) (2, 0.42) (3, 0.33) (4, 0.19)};
        \addlegendentry{Tree of Thoughts}

        \addplot[TOT, mark=*, mark options={fill=TOT}] coordinates {(0, 0.68) (1, 0.62) (2, 0.56) (3, 0.46) (4, 0.32)};
        \addlegendentry{Thread of Thought}

        \addplot[STANDARD, mark=*, mark options={fill=STANDARD}] coordinates {(0, 0.70) (1, 0.67) (2, 0.65) (3, 0.60) (4, 0.52)};
        \addlegendentry{Standard}

        \addplot[S2A, mark=*, mark options={fill=S2A}] coordinates {(0, 0.58) (1, 0.54) (2, 0.49) (3, 0.41) (4, 0.29)};
        \addlegendentry{System 2 Attention}

        \end{axis}
        \end{tikzpicture}
        \caption{\textsc{Economical Prompt Indexes} for each prompt method tested on Llama 3-8B.}
        \label{fig:llama3-8b-line}
    \end{minipage}
\end{figure}

\end{document}